\documentclass[runningheads]{llncs}
\usepackage{amsfonts}
\usepackage{amssymb}
\usepackage{multirow}
\usepackage{cite} 
\usepackage{color}
\usepackage{hyperref}
\usepackage[T1]{fontenc}
\usepackage{graphicx}
\usepackage[table,xcdraw]{xcolor}
\usepackage{multirow}
\begin{document}
\title{MGA:  Medical generalist agent through text-guided knowledge transformation}
\author{Weijian Huang\inst{1,2,3}\and Hao Yang\inst{1,2,3}\and Cheng Li\inst{1}\and Mingtong Dai\inst{1,2,3} \and Rui Yang\inst{4}\and Shanshan Wang\inst{1,3,*}}
\authorrunning{W. Huang et al.}
\titlerunning{Text-guided X-Ray generalist agent}
\institute{Paul C. Lauterbur Research Center for Biomedical Imaging, Shenzhen Institutes of Advanced Technology \and University of Chinese Academy of Sciences, Beijing, China, Chinese Academy of Sciences, Shenzhen, Guangdong, China
\and Pengcheng Laboratory, Shenzhen, Guangdong, China
\and Renmin hospital of wuhan university, Urology department, Wuhan, China
\\
\email{*Corresponding:Sophiasswang@hotmail.com, ss.wang@siat.ac.cn}}

% \author{***}
% \institute{***}
\authorrunning{Weijian Huang et al.}
\titlerunning{Text-guided medical generalist agent}

\maketitle              % typeset the header of the contribution
\begin{abstract}
Multi-modal representation methods have achieved advanced performance in medical applications by extracting more robust features from multi-domain data. However, existing methods usually need to train additional branches for downstream tasks, which may increase the model complexities in clinical applications as well as introduce additional human inductive bias. Besides, very few studies exploit the rich clinical knowledge embedded in clinical daily reports. To this end, we propose a novel medical generalist agent, MGA, that can address three kinds of common clinical tasks via clinical reports knowledge transformation. Unlike the existing methods, MGA can easily adapt to different tasks without specific downstream branches when their corresponding annotations are missing. More importantly, we are the first attempt to use medical professional language guidance as a transmission medium to guide the agent's behavior. The proposed method is implemented on four well-known X-ray open-source datasets, MIMIC-CXR, CheXpert, MIMIC-CXR-JPG, and MIMIC-CXR-MS. Promising results are obtained, which validate the effectiveness of our proposed MGA. Code is available at: \href{https://github.com/SZUHvern/MGA}{\emph{\textcolor[rgb]{0,0,1}{\underline{git.MGA}}}}
% code will be released \emph{\textcolor[rgb]{0,0,1}{\underline{Here}}}.

\keywords{Multi-modal representation \and Image-text representation \and Generalist agent}
\end{abstract}
\section{Introduction}

\begin{figure}
\centering
\includegraphics[width=0.8\textwidth]{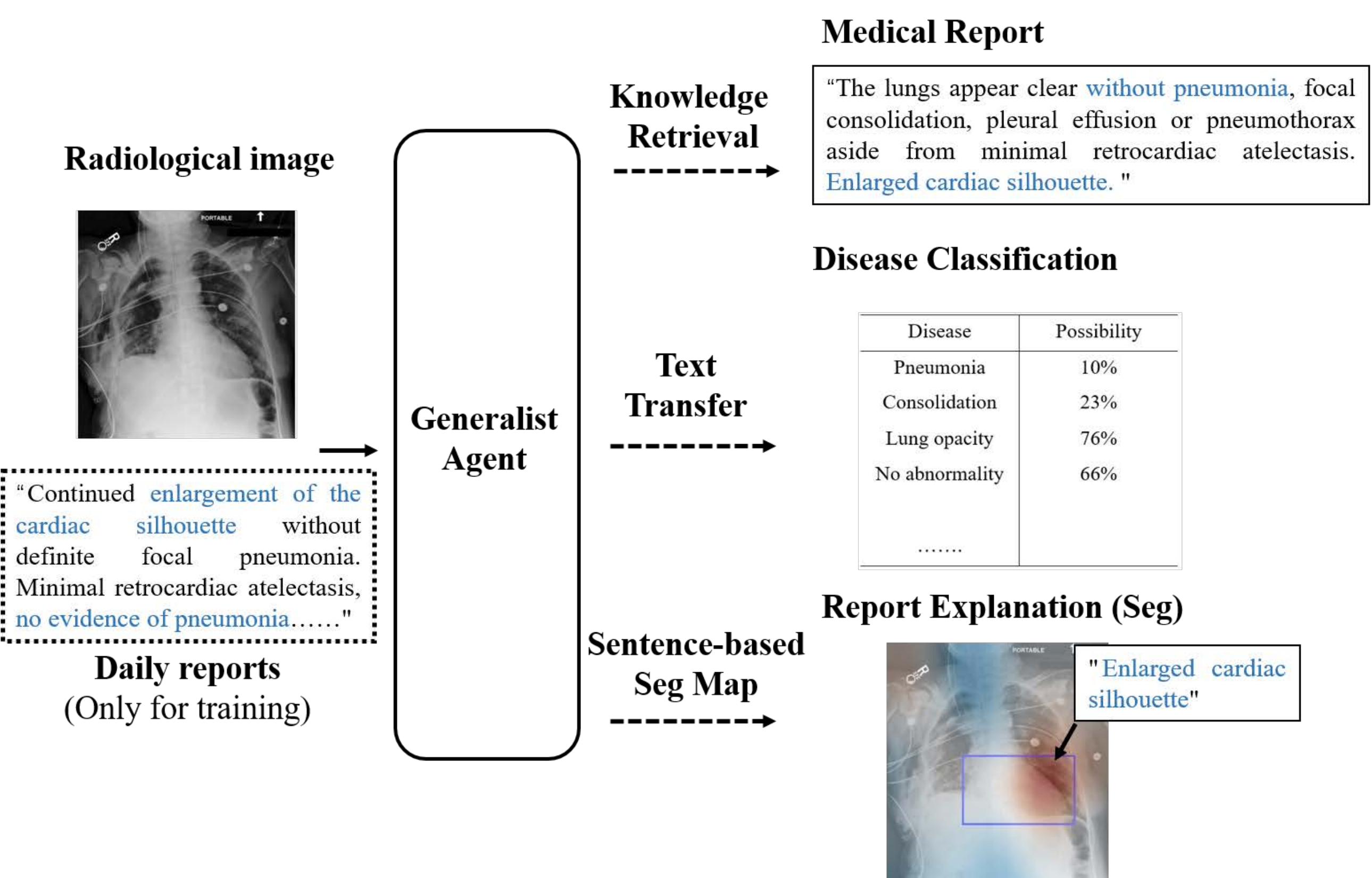}
\caption{Workflow of the proposed Medical Generalist Agent (MGA). Three common clinical tasks (classification, segmentation, and report generation) can be accomplished without training downstream task-specific branches.}
\label{fig1}
\end{figure}
Advances in machine learning have enabled automated diagnostic systems (ADS) to approach expert-level performance, making it feasible to use deep learning to improve the clinical workflow \cite{chilamkurthy2018deep, titano2018automated}. Common clinical applications, including disease diagnosis, lesion quantification, and medical report generation, can be properly handled by training different machine learning methods. Single task-based ADS methods \cite{wang2021annotation, zhou2019d, huang2021coarse} can efficiently address respective tasks individually. These methods obtain high performance by training with a large number of labeled data. Nevertheless, training different models for different applications can be computationally expensive. Moreover, fully supervised methods, which are the current most prevalent practices, can be subject to severe inductive biases due to the annotation process \cite{reed2022generalist}.

On the other hand, by extracting more robust features from multi-source data, multi-modal methods \cite{zhou2022generalized, boecking2022making, moon2022multi, Lee, tiu2022expert} have achieved state-of-the-art performance in many medical tasks. One typical example of multi-source data is paired clinical reports and X-ray images \cite{dunnmon2020cross, irvin2019chexpert, johnson2019mimic}. The reports are written by qualified physicians summarizing his/her findings for disease judgment after inspecting the images. Both reports and images contain important information that can assist downstream tasks, such as disease diagnosis and lesion segmentation. A unified model, which is trained using these multi-source data, can easily adapt to different downstream tasks and make comprehensive decisions. Training a unified model to make different predictions resemble more clinical practice. Therefore, multi-modal methods show a promising direction for the next generation of ADS.

However, existing methods are still limited in the following aspects: 1) Model redundancy caused by multiple branches or multiple downstream heads. They tend to treat the unified model as a feature extractor to extract common features, and add different network branches for the specific tasks \cite{wu2023medklip,wang2023chatcad, miuraetal2021, singh2019chest, wang2018tienet, huang2021gloria}. For example, there are works using a large language model (i.e. Chatgpt, Bert \cite{devlin2018bert}) or a recurrent neural network (RNN) to obtain the text prediction, adding several fully connected layers to obtain the targeted classification results\cite{huang2021gloria} or cascading a vision decoder to predict segmentation maps. Although this type of method can easily achieve high performance on specific tasks through the added fully supervised branches, it also increases the computational cost significantly. Moreover, the generalization capability is questionable. 2) The need to manually annotate a large number of data for each clinical task. Data annotation is quite subjective, which may introduce severe human inductive biases. For example, diagnostic report semantics are sometimes inconsistent with the annotated classification labels \cite{johnson2019mimicjpg}. 3) Inadequate use of clinical knowledge. Taking the task of medical report generation as an example. Medical knowledge is highly specialized. Medical reports use a lot of negative texts that are uncommon in natural languages. As a result, it is difficult to generate medical reports by using common natural language-based models (i.e. Bert, GPT\cite{brown2020language}).

In this paper, we propose a novel Medical Generic Agent (MGA, Fig. \ref{fig1}), which can handle the three most common medical tasks without requiring any additional task-specific branches. MGA treats language as a medium to guide the output generation of downstream tasks through the retrieval of medical expert knowledge. The major contributions of this work are listed as follows:

1) To the best of our knowledge, we make the first attempt to aggregate a variety of common medical downstream tasks by proposing a medical generalist agent, MGA. As the main medium of human communication, we choose appropriate language representation to guide the behavior of the agent.

2) MGA is applicable to handling various challenges for common medical tasks, such as missing task-specific annotations or shortage of computing resources. Each of the tasks is aligned by the corresponding designed training strategy. The agent is trained cooperatively, without separate training stages.

3) The method is implemented on four well-known X-ray datasets, MIMIC-CXR\cite{johnson2019mimic}, CheXpert\cite{irvin2019chexpert}, MIMIC-CXR-JPG\cite{johnson2019mimicjpg}, and MIMIC-CXR-MS\cite{boecking2022making}, and competitive results have been achieved when compared with fully-supervised learning methods. 

\section{Method}

The goal of our work is to jointly learn the representation from image-report X-Ray data and to address different tasks without adding specific branches. To achieve this goal, we need to find a medium that can transform in the agent and guide the network action. As the main way of human communication, we believe that language is flexible and effective enough. By decorating language and using its representation in appropriate ways, we establish a promising medical generalist agent. 

\subsection{Image-Language Representation Learning}
\begin{figure}
\centering
\includegraphics[width=0.8\textwidth]{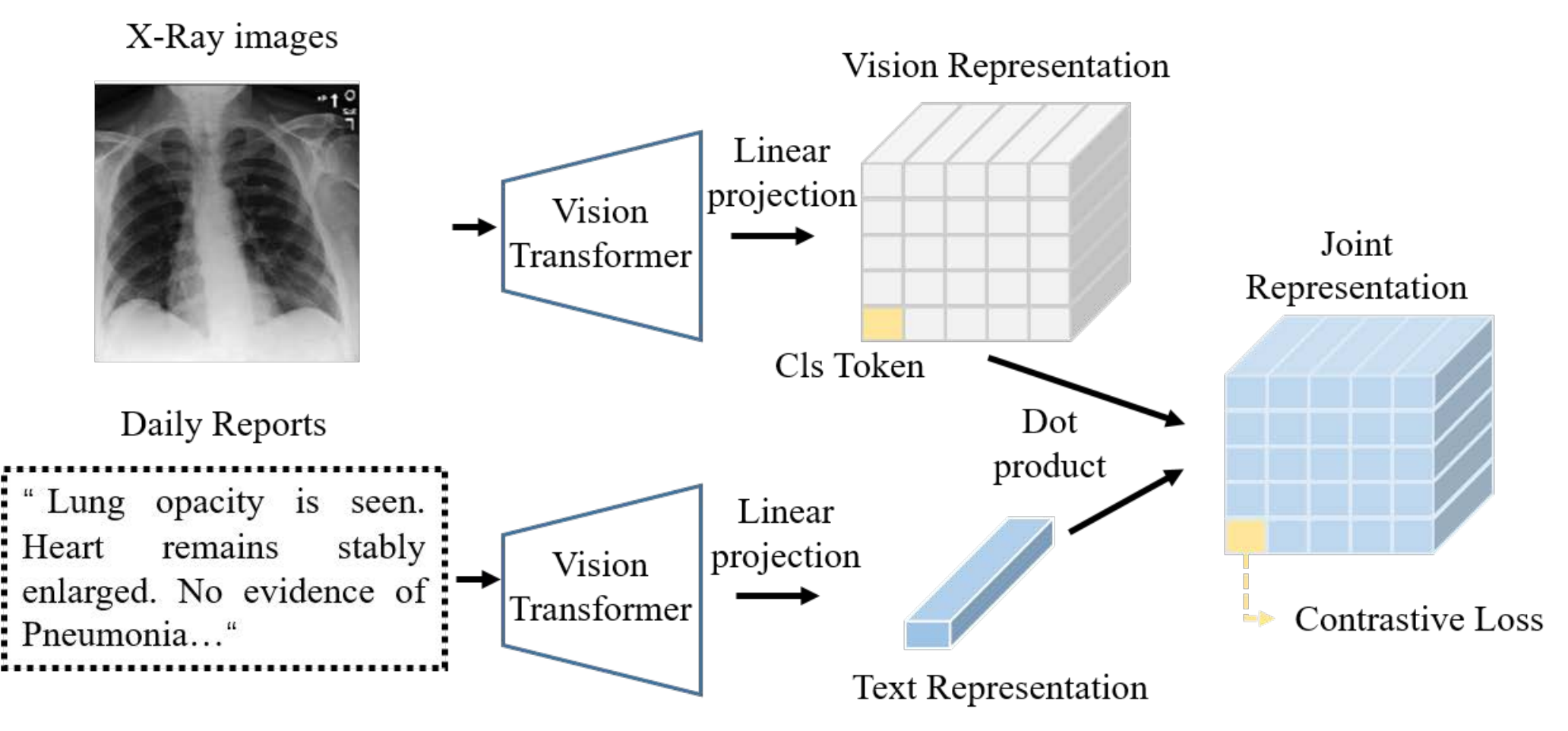}
\caption{The base architecture. MGA adopts CLIP as the baseline and controls the representation of the output by language-guided knowledge transformation.}
\label{fig2}
\end{figure}

CLIP\cite{radford2021learning} is one of the most famous image-text representation methods. In CLIP, two separate encoders are jointly trained with pairwise cosine similarities constraining visual and textual representation learning. By training with a very large scale of paired image-text data, CLIP obtained state-of-art visual-language representation. The workflow of CLIP is shown in Fig. \ref{fig2}. Inspired by CLIP, we proposed three text-guided strategies to achieve our medical generalist agent (Fig. \ref{fig3}). The details will be described in the following sections.

\subsection{MGA for Classification}
\begin{figure}
\centering
\includegraphics[width=1.0\textwidth]{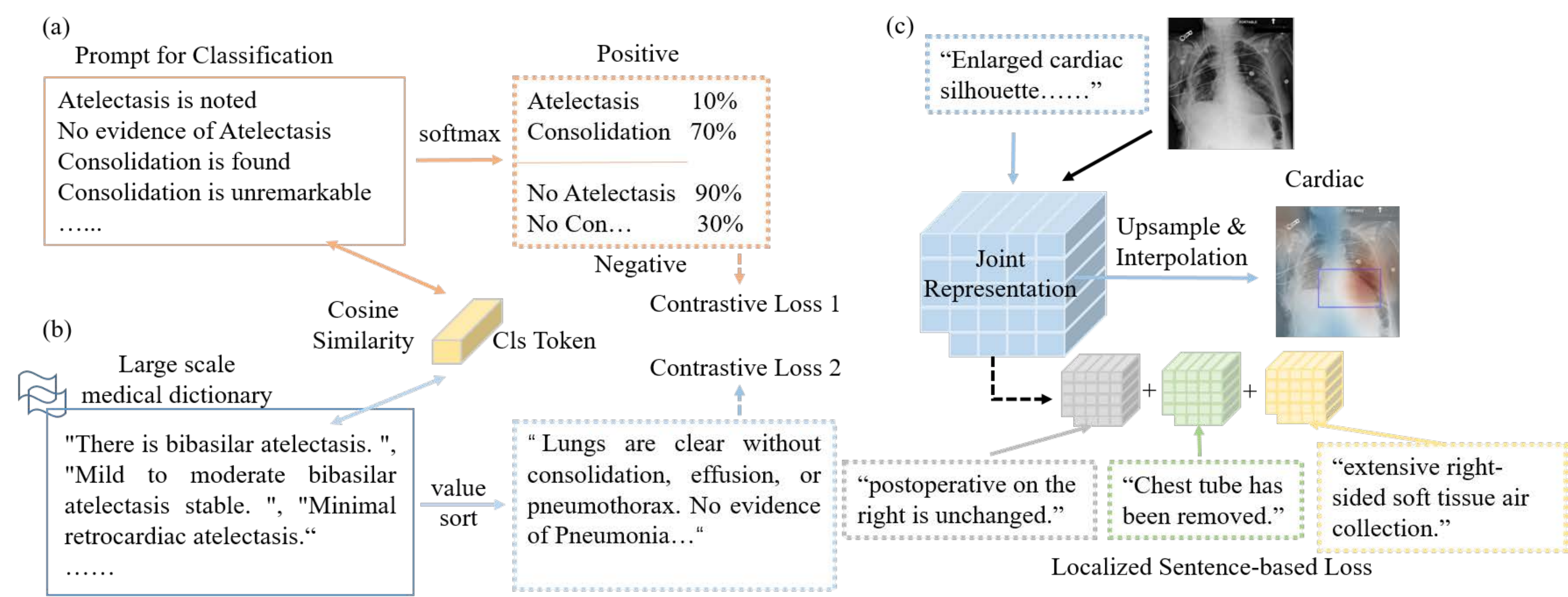}
\caption{Text-guided knowledge transformation to address three common medical application tasks. (a) MGA for classification. Both positive and negative sentences are considered. (b) MGA for report generation. We construct a large medical dictionary by class clustering and select the top-5 similarity score sentences to form the final reports. (c) MGA for segmentation. Sentence-image alignment is achieved by the designed alignment loss. We treat the aligned heatmap as segmentation results.}
\label{fig3}
\end{figure}

Recognizing the disease category according to X-Ray images is a common task in clinical practice. The main challenges of this task include the confusion caused by a large number of negative samples and the difficulty of multi-category differentiation. To address these challenges, we design a list of Intra/Inter class prompts to train and evaluate the agent (Fig. \ref{fig3} (a)):
1) Intra-class prompts should include both positive and negative disease descriptions to make full use of the negative samples in the dataset.
2) Sentences of different categories should be orthogonal in the embedding space to allow agents to implicitly explore the intrinsic relationships between different categories.

The agent is provided with sentences containing paired pos/neg disease information. The representation for each category is calculated with softmax. Thus, the multi-label binary cross-entropy loss function can be defined as: 
\begin{equation}
\label{Eq.3}
Loss_{cls}= \frac{1}{NC}\sum_{i}^{N}\sum_{c}^{C}y_{i,c}log(p_{i,c}) + (1-y_{i,c})log(1-p_{i,c}),
\end{equation}

where $y_{i,c}$ is the binary label of category $c$ in the batch of $i^{th}$ sample, $p_{c}=\sigma(\langle v\cdot t_{pc} \rangle, \langle v\cdot t_{nc} \rangle)$ represents the probability of class $c$, where $\sigma\ (\cdot)$ is the softmax function, $t_{pi}$ and $t_{ni}$  indicates pos/neg sentence representation respectively.

\subsection{MGA for Report Generation}

A contrastive constraint is employed as the basic loss to learn the visual and language joint representation. The loss function is:
\begin{equation}
\label{Eq.1}
Loss_{basic}(v,t)= -\sum_{i}^{N}log(\frac{exp(\langle v_{i},t_{i} \rangle)}{\sum_{k}^{N}exp(\langle v_{i},t_{k}\rangle)})
\end{equation}
where $\langle v_{i},t_{i} \rangle$ represents the cosine similarity between the visual representation $v=[v_{1},...,v_{N}]$ and text representation $t=[t_{1},...,t_{N}]$.

Unlike natural languages, medical reports are written by clinicians. The texts utilized are highly professional and flexible. Common language models are difficult to capture these specialized contexts, thus it is difficult to automatically generate a medical report from scratch. To this end, we choose to match the most relevant sentences according to previous experience and knowledge rather than directly generate medical reports during evaluation. The procedure of our method is shown in Fig. \ref{fig3} (b).

Specifically, we construct a medical dictionary according to retrospective reports. To avoid overlapping sentences with the same semantics, we use K-means to cluster the sentences in the dictionary. Here, the term frequency-inverse document frequency (TF-IDF) is selected to calculate the sentence distance \cite{beel2016paper}. The representation of each word $t$ in a sentence $d$ with length $K$ is define as: 
\begin{equation}
\label{Eq.2}
s_{i} = tf(t_{i},d)\cdot idf(t_{i}, D),
\end{equation}
where term frequency $tf(t,d)=\frac{f_{t,d}}{\Sigma_{t^{'}\in d}tf(t^{'},d)}$ indicates the relative frequency of term $t$ within the sentence $d$. $f_{t,d}$ is the number of times that term $t$ occurs in the sentence $d$. The inverse document frequency $idf(t,D)=log\frac{N}{|d\in D:t\in d|}$ is a measure of how important the word is in the dictionary $D$. Thefore, the representation vector of $d$ can be written as $s_{d}=(s_{d1},..,s_{dk})$.

MGA calculates the similarity in the dictionary and selects the sentences with the highest similarity from different clusters to form the final report. The obtained reports are more flexible rather than those generated by template-matching approaches.

\subsection{MGA for Segmentation}
Since it is difficult to superimpose pixel-level supervision on downsampled images (MGA does not have any up-sampling decoder branch), we use language with semantic information as an auxiliary constraint. We establish a patch-level relationship between the sentence and the image and treat the similarity matrix as the segmentation result (Fig. \ref{fig3} (c)). 

In detail, an image patch is determined according to the downsampling level of the visual encoder while the sentence is extracted from the medical reports during training. Here, we use an alignment strategy. Unlike GloRIA\cite{huang2021gloria}, which uses fixed word alignment, we treat the sentence as the basic unit since it is difficult to represent medical semantics in one word. For the report that cannot meet the number of sentences at the training stage (a fixed number to calculate the attention map), we pad blank text representation and ignore the corresponding calculation when performing backpropagation. 

% Assuming $s_{ij}\in \mathbb{R}^{P\times T}$ is the similarity between the representation of patch $j$ in the image and sentence $i$ in the report. A sentences-patches weighted map $c_{i}\in \mathbb{R}^{P\times T}$ can be written as:
% \begin{equation}
% \label{Eq.4}
% c_{i} = \sum_{j}^{P}\frac{exp(s_{ij})}{\sum_{k}^{P}exp(s_{ik})}v_{j}
% \end{equation}

 Assuming the localized feature matching function $Z(x_{ti}, x_{vi})$ to aggregate the similarities between all sentence features and their corresponding attention-weighted image features. Similar to \cite{huang2021gloria}, $x_{ti}$ and $x_{vi}$ are local features for the $i$th report and image. The localized sentence-based constraint $Loss_{seg}$ is then define as:

\begin{equation}
\label{Eq.4}
Loss_{seg}(x_{v},x_{t})= -\sum_{i}^{N}log(\frac{exp(Z(x_{ti},x_{vi}))}{\sum_{k}^{N}exp(Z(x_{tk},x_{vi}))})
\end{equation}
The final training loss for our MGA is:
\begin{equation}
\label{Eq.5}
Loss_{all} = \tau_{1}Loss_{basic} + \tau_{2}Loss_{cls} +  \tau_{3}Loss_{seg}
\end{equation}
where $\tau$ is a hyper-parameter.

\section{Experiments and Results}

\subsection{Datasets}
Four well-known chest X-ray datasets are adopted:
1)\textbf{MIMIC Chest X-ray v2}\cite{johnson2019mimic} (MIMIC-CXR) - A large dataset of chest radiographs with free-text radiology reports. 377,110 images corresponding to 227,835 radiographic studies were collected.
2)\textbf{CheXpert}\cite{irvin2019chexpert} - Retrospectively collected chest radiographic examinations from Stanford Hospital. 224,316 chest radiographs of 65,240 patients with high-quality classification labels are provided.
3)\textbf{MIMIC-CXR-JPG}\cite{johnson2019mimicjpg} - A dataset derived from MIMIC-CXR. Structured labels extracted from the free-text reports are provided. A standard reference for data splits and image labels is given. Particularly, an open-source-based language tool, NegBio, is used to extract classification labels from the free-text reports. We use this dataset to split training and testing sets, as well as to evaluate the classification capability of MGA.
4)\textbf{MIMIC-CXR-MS}\cite{boecking2022making} - Another dataset derived from MIMIC-CXR. 1162 cases from MIMIC-CXR are provided with bounding boxes for disease detection. We use this dataset only to evaluate the segmentation capability of MGA.

In order to fully extract the relationship between language and images, we enlarge the training set by combining MIMIC-CXR with CheXpert. For CheXpert, we use prompt\cite{radford2021learning} to generate radiographs according to the class labels. We evaluate the method on CheXpert, MIMIC-CXR-JPG, and MIMIC-CXR-MS. We make sure that there is no data leakage issue, i.e., no case appears simultaneously in the training and testing sets.

\subsection{Implementation Details}
Only frontal view scans are used in all experiments. In total, there are 4,411,811 training images. All comparison algorithms are implemented with the same data preprocessing procedure on the same program platform to ensure the fairness of comparison. For the comparison CNN models (Resnet50 and Densenet121), we followed PyTorch's classical setting with pre-training weights, and only the final full connection layer is modified to adapt to the specific classification task. The visual encoder and text encoder of MGA are the same as those of CLIP\cite{radford2021learning}. We use Adam optimizer with a learning rate of $1\times 10^{-5}$. For the number of sentences needed for the segmentation constraint, we set it to 5 by default. 

\subsection{Disease Classification}
The classification performance of MGA is investigated in this section. Following the setting of CheXpert \cite{irvin2019chexpert}, 8 types of diseases are selected to calculate the classification metrics. Specifically, the metrics of individual diseases are calculated and the averaged results are reported. We train MGA and the baseline CLIP on MIMIC-CXR dataset with free reports while training Resnet and Densenet on MIMIC-CXR-JPG with one-shot classification labels. Results are listed in Table \ref{table1}.

\begin{table}[!ht]
\caption{Classification results of different methods. The methods are trained on MIMIC-CXR and evaluated on MIMIC-MS and CheXpert.}
\centering
\setlength{\tabcolsep}{1.8mm}{
\begin{tabular}{|c|ccc|ccc|}
\hline
\multirow{2}{*}{Methods} & \multicolumn{3}{c|}{MIMIC-MS}  & \multicolumn{3}{c|}{CheXpert}  \\ \cline{2-7} 
                         & Acc   & F1    & AUC            & Acc   & F1             & AUC   \\ \hline
Resnet(class-level)      & 79.72 & 81.25 & \textbf{79.95} & 74.62 & 70.17          & 81.67 \\
DenseNet(class-level)    & 77.39 & 79.40 & 78.85          & 75.75 & \textbf{71.37} & 83.81 \\
CLIP(text-level)         & 57.25 & 63.12 & 64.31          & 67.51 & 67.52          & 78.15 \\
MGA(text-level) & \textbf{86.48} & \textbf{81.51} & 71.01 & \multicolumn{1}{l}{\textbf{75.06}} & \multicolumn{1}{l}{69.91} & \multicolumn{1}{l|}{\textbf{82.32}} \\ \hline
\label{table1}
\end{tabular}}
\end{table}

Results show that MGA performs better than CLIP when evaluated on both datasets. It is also encouraging to see that even without using any classification head, overall, our proposed MGA can achieve higher classification accuracy than the classic deep learning methods, indicating its strong capability in visual-language representation learning.

\subsection{Reports Generation}
\begin{figure}
\centering
\includegraphics[width=0.9\textwidth]{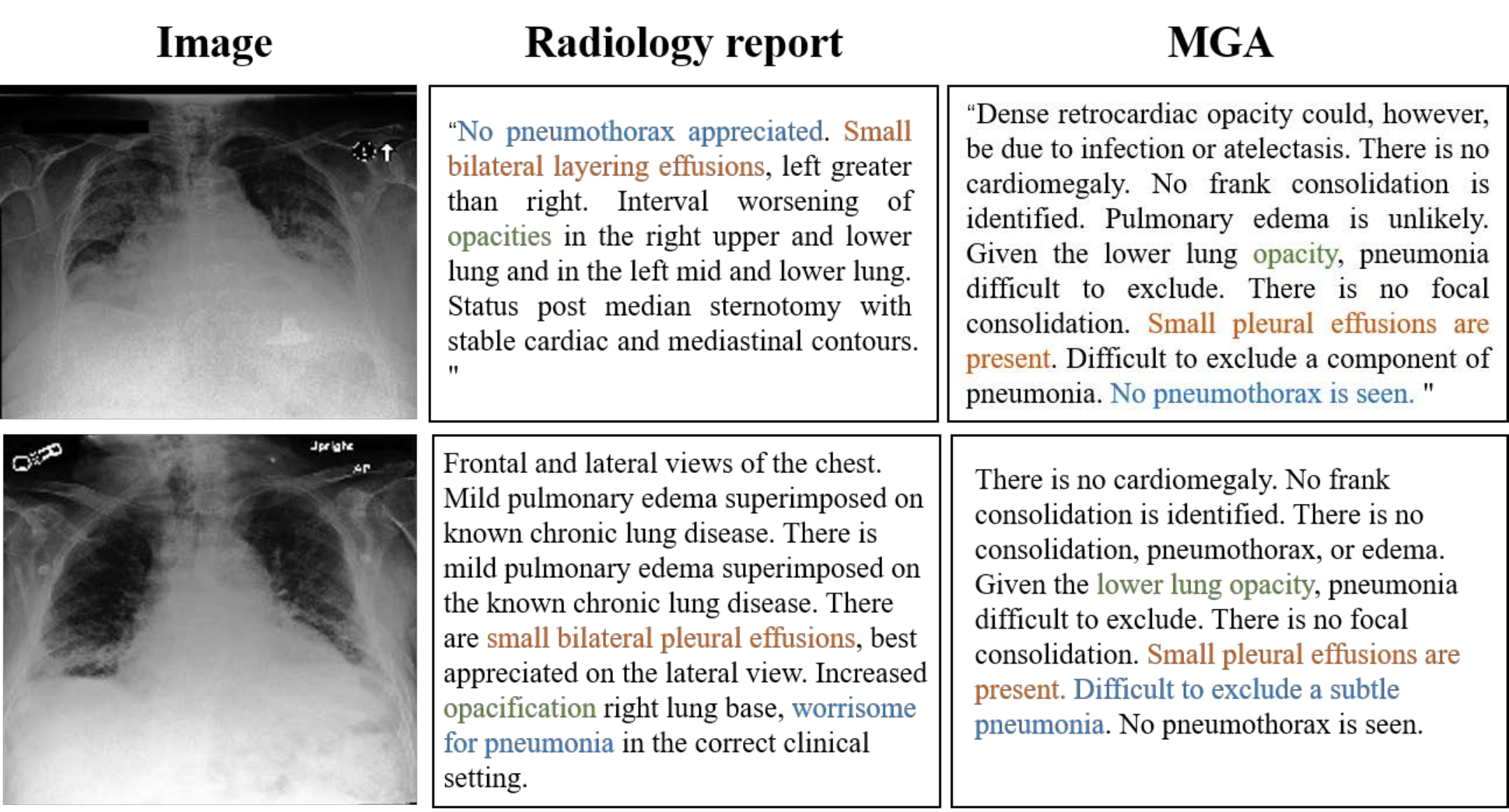}
\caption{Two examples of medical report generation. Reports generated by MGA are compared to those provided by radiologists. The blue color shows identical sentences. The orange color indicates sentences with the same semantic information.}
\label{fig4}
\end{figure}

The generated reports are as shown in Fig. \ref{fig4}. We color the generated sentences in the reports in a similar way to those of radiologists. Although there are some expression differences, we find that the generated sentences resemble very much those given by radiologists. This observation validates that MGA has learned the correct relationship between images and medical reports. More importantly, MGA exhibits the capability of extracting semantic information from sentences.

\subsection{Lesion Segmentation}
\begin{figure}
\centering
\includegraphics[width=1.0\textwidth]{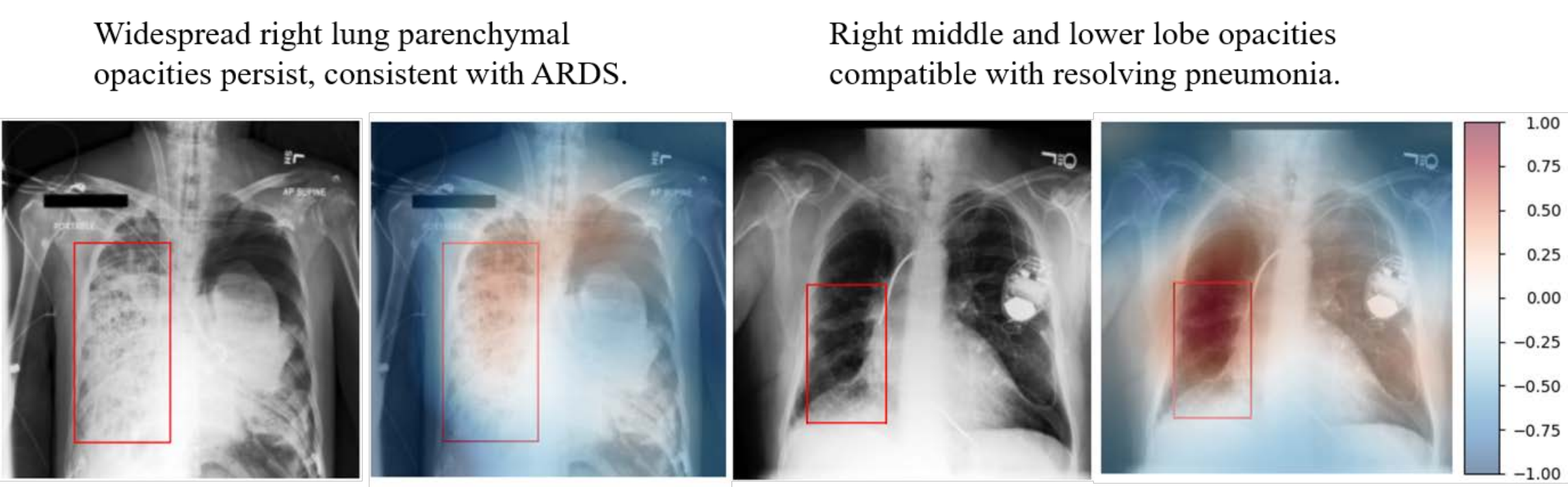}
\caption{Example segmentation results. The red bounding boxes show ground truth references corresponding to the text descriptions while the heatmaps are the segmentation results generated by MGA.}
\label{fig5}
\end{figure}

Finally, we evaluate the segmentation performance of MGA on MIMIC-CXR-MS (Fig. \ref{fig5}). We calculate the similarity between text representation and image-patch representation one by one. Then, the segmentation heatmap is generated through bilinear up-sampling. Satisfactory overlapping between the responses of image and text representation is observed, which reflects the feasibility of employing MGA for lesion segmentation. Nevertheless, there is still misalignment that should be improved in future developments.

\section{Conclusion}

In this work, we propose a medical generalist agent (MGA), which can be generalized to most clinical tasks via knowledge transformation without any task-specific branches/heads. Extensive experiments on four well-known chest X-ray datasets have shown that MGA is applicable to handling various challenges, such as missing task-specific annotations or shortage of computing resources. Three common clinical application tasks (disease classification, report generation, and lesion segmentation) are investigated, and MGA consistently shows very inspiring performances. MGA can be further improved by decorating with the specific task heads, but this is beyond the scope of this study. In conclusion, we have shown a promising direction of the medical generalist agent, which has a strong task generalization ability while relieving the need for task-specific annotation. 

% \subsubsection{Acknowledgments}
% This research was partly supported by Scientific and Technical Innovation 2030-"New Generation Artificial Intelligence" Project (2020AAA0104100, 2020AAA0104105), the National Natural Science Foundation of China (62222118, U22A2040), Guangdong Provincial Key Laboratory of Artificial Intelligence in Medical Image Analysis and Application (2022B1212010011), the Basic Research Program of Shenzhen (JCYJ20180507182400762), Shenzhen Science and Technology Program (RCYX20210706092104034), and Youth Innovation Promotion Association Program of Chinese Academy of Sciences (2019351).

% ---- Bibliography ----
%
% BibTeX users should specify bibliography style 'splncs04'.
% References will then be sorted and formatted in the correct style.
%
\clearpage
\bibliographystyle{splncs04}
\bibliography{mybibliography}
\end{document}